# Spot: A Natural Language Interface for Geospatial Searches in OSM

Lynn Khellaf[1,*] Ipek Baris Schlicht[1], Julia Bayer[1], Ruben Bouwmeester[1], Tilman Miraß[1] and Tilman Wagner[1]

[1] Research and Cooperation Projects, Deutsche Welle, Bonn, Germany; lynn.khellaf@dw.com, ipek.baris-schlicht@dw.com, julia.bayer@dw.com, ruben.bouwmeester@dw.com, tilman.mirass@dw.com, tilman.wagner@dw.com

* Author to whom correspondence should be addressed.

This abstract was accepted to the OSM Science 2023 Conference after peer-review.

Investigative journalists and fact-checkers have found OpenStreetMap (OSM) [1] to be an invaluable resource for their work due to its extensive coverage and intricate details of various locations, which play a crucial role in investigating news scenes. However, the accessibility and usability of this tool pose significant challenges for individuals without a technical background. This need for simplified access to OSM data has brought attention to the potential of Large Language Models (LLMs). Known to the public through applications like OpenAI's ChatGPT [2], Meta's Llama [3], or Google's Bard [4], these models have showcased remarkable capabilities in tackling a range of natural language processing (NLP) tasks. Nevertheless, their deployment in the context of low-resourced languages, both natural and for programming, comes with its own set of challenges, as is the case with OSM query languages such as Overpass Turbo (OT) [5] - an essential OSM data querying tool.

Notably, individuals have started using systems like ChatGPT to formulate OT queries due to the complexity of the language, highlighting the desire for easier access to OSM data. However, the results of such attempts are often only half-convincing. A critical issue has emerged with these models' tendency to hallucinate, leading to the generation of erroneous and non-executable outputs. Other prior efforts [6, 7] were highly limited in their functionality and performance, further underscoring the need for effective solutions in this domain.

The central focus in our KID2 ("KI gegen Desinformation #2") project is therefore to enable a broader audience to seamlessly query the extensive OSM database without the prerequisite of an in-depth understanding of the intricate OSM tagging system or complex OSM query languages like Overpass Turbo. This is achieved through the development of Spot, a robust natural language interface tailored explicitly for querying OSM data, finding Spots, combinations of objects in the public space. Users' natural sentences are translated by a transformer model and turned into OSM database searches. The primary use case within the KID2-project is geo-location in the context of verification, a capability important for (investigative) journalists who often require precise location information for their work.

The application includes a user-friendly graphical interface where users can effortlessly enter their textual search requests, with the subsequent results being visually displayed on the map. This system architecture takes inspiration from the existing Overpass Turbo; however, a conscious decision was made to develop proprietary database query methods to enhance flexibility and functionality.

To construct the foundation for this interface, the OSM data was transformed into a database format that allows for fast and easy data access. This included the selection and processing of a subset of OSM tags – specifically, those that are both visible and instrumental in describing a scene, resulting in a 20% reduction in data size. The subsequent step involved importing this processed dataset into a Postgres (https://www.postgresql.org/) instance. Additionally, the Postgres instance was enhanced with the PostGIS [8] extension to facilitate geospatial queries.

To align with the overarching goal of ensuring accessibility for a wider user base unacquainted with the complexities of OSM tagging, visually similar tags were grouped logically, and the bundles were assigned numerous natural words users might use to describe the corresponding objects. The result is a novel table structure containing information from the OSM database regarding the tag bundles, the natural descriptors, and a list of tags that tend to appear in combination with the current bundle frequently, as extracted from the OSM database.

As part of our training data generation process, we randomly generated tag samples representing objects from the underlying database to construct artificial queries. Each drawn object can be assigned multiple tags based on the previously extracted frequent tag combinations. This process was iterated multiple times to generate an extensive list of random tag combinations, each representing one artificial user query. The selected tags were then stored in a novel intermediate graph-database format. Aside from an area specification (area name or blank if area is defined in the UI), this format encompasses objects with a variable number of tags as nodes, while the edges consist of distance information between the objects where required. This database laid the groundwork for generating effective OSM database queries.

Harnessing the power of ChatGPT, tailored prompts were generated using the selected tag combinations. These prompts were used to instruct ChatGPT to generate natural search sentences emulating user queries. Aside from all the query information from the intermediate format, style instructions were given to improve language variation. The subsequent step focused on training a neural network to translate these natural sentences back into the novel graph-database format. The chosen translation model was a text-to-text transformer which is pretrained on Common Crawl (https://commoncrawl.org/) and a variety of supervised task datasets, T5 [9]. The decision to develop a custom machine learning model was due to the necessity to have full control over bias, accuracy and data privacy. Future tests might include the use of open-source LLMs as the translation model. Figure 1 provides a visual representation of the sequential steps within the pipeline through an example sentence.

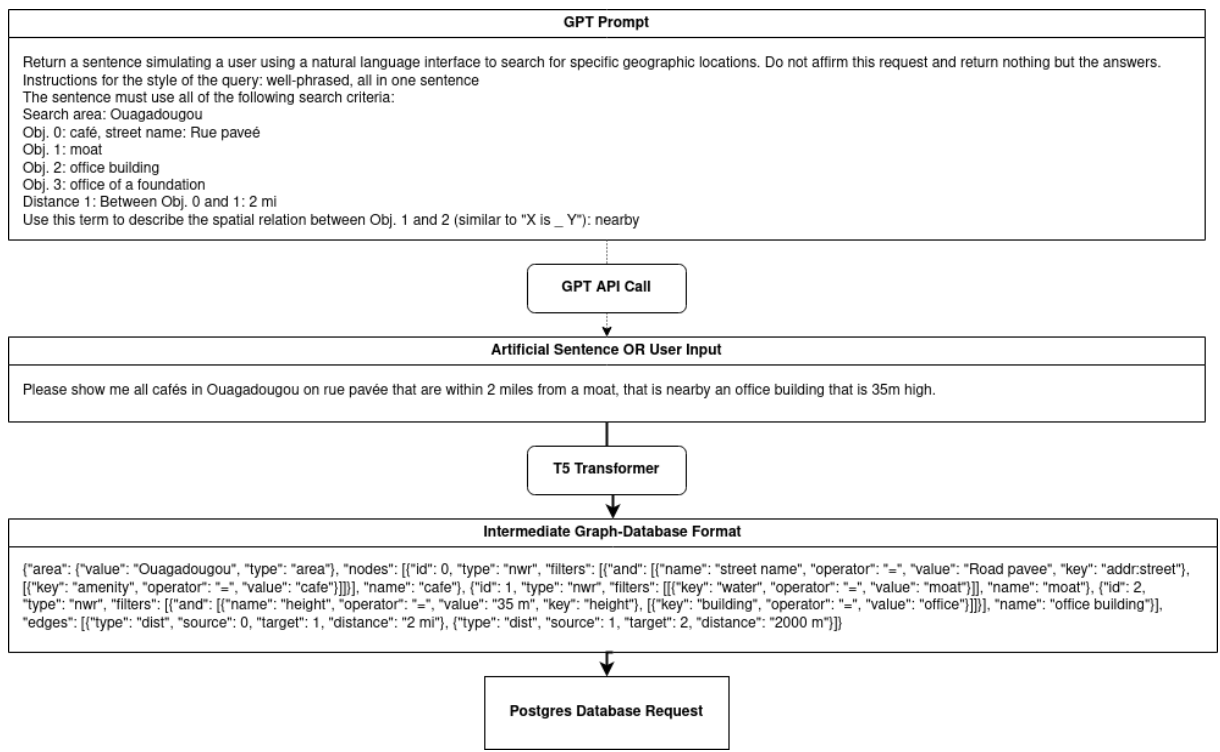

*Figure 1*. Illustration of the pipeline that transforms a natural sentence into a PostGres request to search the OSM database. The first container shows a GPT prompt that is only used during the generation of artificial training data.

Evaluation criteria of the model output are the validity of the format and the semantic accuracy of the extracted information. Benchmarking, which is going to be performed in the near future, will involve a comparison against established natural language OSM interfaces and common LLMs like GPT-4 [10]. To achieve this objective, we are currently developing a comprehensive, gold-standard natural language query dataset that encompasses all the desired use cases relevant to geolocation verification.

The current project goal is to construct a functional prototype for use in investigative journalism with a broad, but finite set of possible query structures. A first working proof-of-concept exists and is currently undergoing a process of iterative improvements. Future versions can be scaled up for different use cases by adapting the underlying database and prompting templates. Additionally, to foster collaboration and future advancement, all code and generated data is available as an open source repository under the following link: https://github.com/dw-innovation/kid2-spot. A public beta release of a usable demo is planned for February 2024.

In summary, we think this research could mark a further step towards democratizing access to OSM data through the development of a cutting-edge natural language interface. By bridging the gap between intricate query languages and user-friendly interactions, this interface has the potential to greatly improve how investigative journalists and a broader audience interact with geospatial data.

### Acknowledgments

This project is led by the Deutsche Welle Research and Cooperation Projects teams and was co-funded by BKM ("Beauftragte der Bundesregierung für Kultur und Medien," the German Government's Commissioner for Culture and Media).

Map data copyrighted OpenStreetMap contributors and available from https://www.openstreetmap.org.